\documentclass[conference]{IEEEtran}
\ifCLASSINFOpdf
\else
\fi
\usepackage{graphicx}
\graphicspath{{images/}}
\usepackage[rightcaption]{sidecap}

\usepackage{wrapfig}
\usepackage{verbatim}
\usepackage{hyperref}
\usepackage{caption}
\usepackage{subcaption}
\usepackage{todonotes}
\usepackage{multirow}
\usepackage{amsmath}

\newcommand{\shortsection}[1]{\vspace*{1ex}\noindent{\bf #1.}}

\begin{document}
%
\title{Context-aware Monitoring in Robotic Surgery}

\author{\IEEEauthorblockN{Mohammad Samin Yasar, David Evans, Homa Alemzadeh}
\IEEEauthorblockA{
University of Virginia, Charlottesville, Virginia\\ 
\{msy9an, evans, alemzadeh\}@virginia.edu}
}


%


\maketitle
\thispagestyle{plain}
\pagestyle{plain}

\begin{abstract}
Robotic-assisted minimally invasive surgery (MIS) has enabled procedures with increased precision and dexterity, but surgical robots are still open loop and require surgeons to work with a tele-operation console providing only limited visual feedback. In this setting, mechanical failures, software faults, or human errors might lead to adverse events resulting in patient complications or fatalities. We argue that 
impending adverse events could be detected and mitigated by applying context-specific safety constraints on the motions of the robot. We present a context-aware safety monitoring system which segments a surgical task into subtasks using kinematics data and monitors safety constraints specific to each subtask. To test our hypothesis about context specificity of safety constraints, we analyze recorded demonstrations of dry-lab surgical tasks collected from the JIGSAWS database as well as from experiments we conducted on a Raven II surgical robot. Analysis of the trajectory data shows that each subtask of a given surgical procedure has consistent safety constraints across multiple demonstrations by different subjects. Our preliminary results show that violations of these safety constraints lead to unsafe events, and there is often sufficient time between the constraint violation and the safety-critical event to allow for a corrective action. 

\end{abstract}


%
\IEEEpeerreviewmaketitle

\section{Introduction}
With the increasing adoption of Robotic Surgical Assistants (RSA) such as Intuitive Surgical's da Vinci Systems, MIS has become the standard approach to certain procedures in urology, gynecology, and general specialties. MIS increases precision and dexterity compared to laparoscopy and open surgery by providing 3D magnified views of surgical field and scaling the motions of miniaturized surgical instruments. RSAs are designed with data logging mechanisms which enable offline analysis of both the kinematics and video data from the procedures. 
However, there is at present no way of understanding context and differentiating between safe and unsafe gestures during the procedure. Our goal is to use this data to infer the surgical context during operation and provide feedback to assist surgeons in avoiding adverse events.

\begin {figure}[b]
   \vspace{-1em}
\centering
    \includegraphics[width=0.48\linewidth]{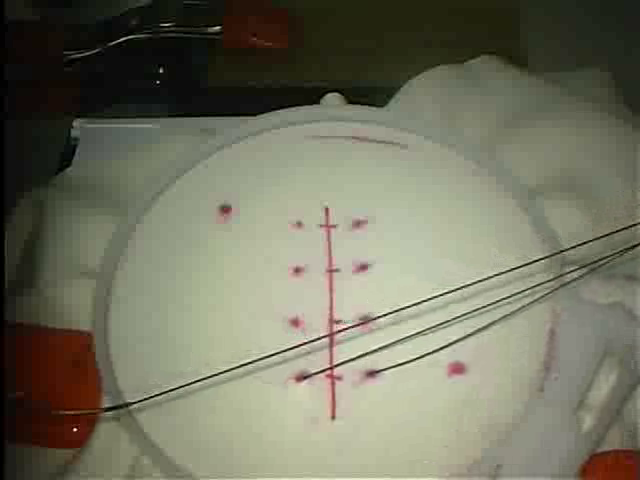}
    \includegraphics[width=0.48\linewidth]{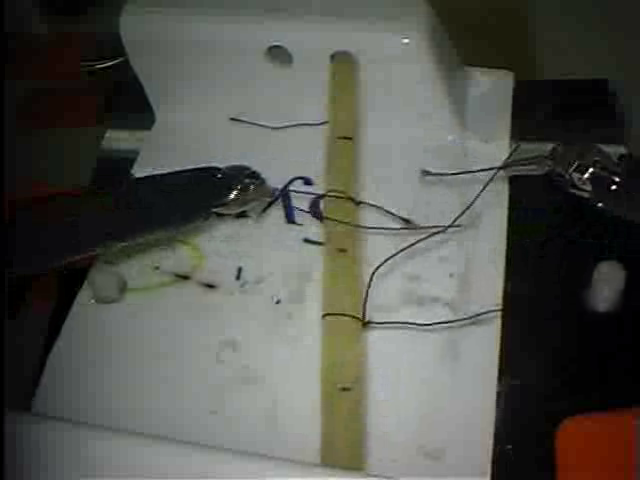}
    \caption{\label{fig:vision_case} Example Cases where Vision Feedback is Inadequate (from JIGSAWS Database \cite{gao2014jhu}).}
\end{figure} 

\begin {figure}[b]
   \vspace{-1em}
\begin{center}
    \includegraphics[width=0.98\linewidth]{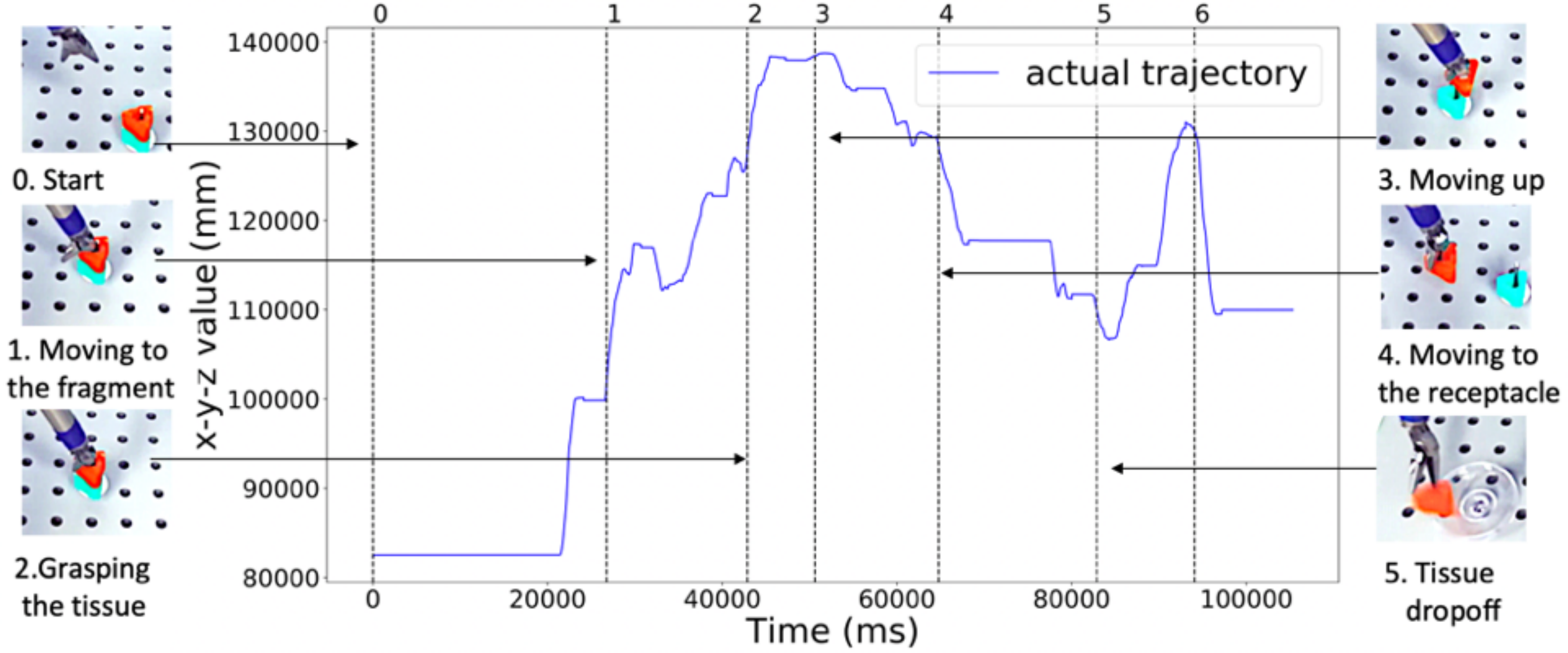}
    \caption{Changes in Constraints across a Surgical Trajectory. The value plotted is the Euclidean distance from the origin to the robot end-effector.}
    \label{fig:rq2_2}
   \vspace{-1em}
\end{center}
\end{figure}

Previous work has demonstrated that the safety of surgery may be compromised by human errors \cite{eubanks1999objective, elhage2015assessment, joice1998errors} or faults in the surgical robots that cause safety-critical events during surgery, such as unintended movements and collision of surgical instruments, modification of surgeon's intent, and unresponsive robotic systems \cite{alemzadeh2014systems}. Examples of technical failures include disruptions of the communication between the surgeon console and the robot, causing packet drops or delays in tele-operation~\cite{bonaci2015make}, and accidental or malicious faults in the robot control software~\cite{alemzadeh2016targeted}. Such adverse events can potentially harm the patients, causing unexpected cuts, bleeding, or minor injuries that further lead to complications during the procedure or afterwards~\cite{alemzadeh2016adverse}. 


Safety in surgery is considered as an intrinsic property of the procedure itself and is often left to the surgeons’ knowledge about the task and the surgical system. 
However, the current open-loop tele-operation setup for MIS provides only video feedback which is inadequate for timely anticipation of adverse events. Figure~\ref{fig:vision_case} shows two examples from the JIGSAWS database~\cite{gao2014jhu} where poor point-of-view or occlusion causes difficulty in precisely monitoring instrument motions, potentially leading to errors such as wrong site injury and unintended applied force among others.

Our analysis and simulation of several tasks from the JIGSAWS database suggests that (1) each surgical task has a specific pattern that is loosely followed by every surgeon performing the task, and (2) each subtask in a particular surgical task has specific parametric constraints which, if violated, can lead to safety-critical events. Figure~\ref{fig:rq2_2} shows the trajectory of the robot end-effector in Cartesian space ($x,y,z$ values) for an example surgical task (debridement); annotated with its subtasks and parametric constraints (Cartesian position of the end-effector).  The goal of our work is to explore the possibility of assisting a surgeon operating an RSA by generating a warning when a substask-specific safety-critical parametric constraint is violated. As a first step, this work evaluates the potential for constraint violations to be used to reliably trigger alerts to the surgical team in advance of a failure.  Although we anticipate these alerts being incorporated into an RSA, in this paper we do not consider or evaluate different options for responding to a detected violation. 


\par
\shortsection{Contributions}
This paper presents our design of a context-aware monitoring system that identifies the current subtask of a surgical task and observes violations of context-specific safety constraints (Section~\ref{sec:context}). The proposed monitoring system involves a learning phase, where subtasks and associated constraints are learned, and a monitoring phase, where violations of constraints are detected during surgery. Our detection only focuses on the analysis of kinematics features during surgery to overcome scenarios where vision feedback might be compromised due to occlusion, poor image quality, or sub-optimal point of view. 

We evaluate the effectiveness of our design by analyzing traces collected from debridement surgical task (Section~\ref{sec:experiments}). We use the recorded video data to evaluate the performance of the detection system, which was only permitted to use kinematics data. 
Our results provide evidence that context is necessary for safety monitoring in robotic surgery, and that by using context it is feasible to anticipate likely failures while there is still time for a surgeon to avoid them.


\par

\begin {figure*}[htp!]
\begin{center}
    \includegraphics[width=0.7\linewidth, trim={0 0 0cm 0},clip]{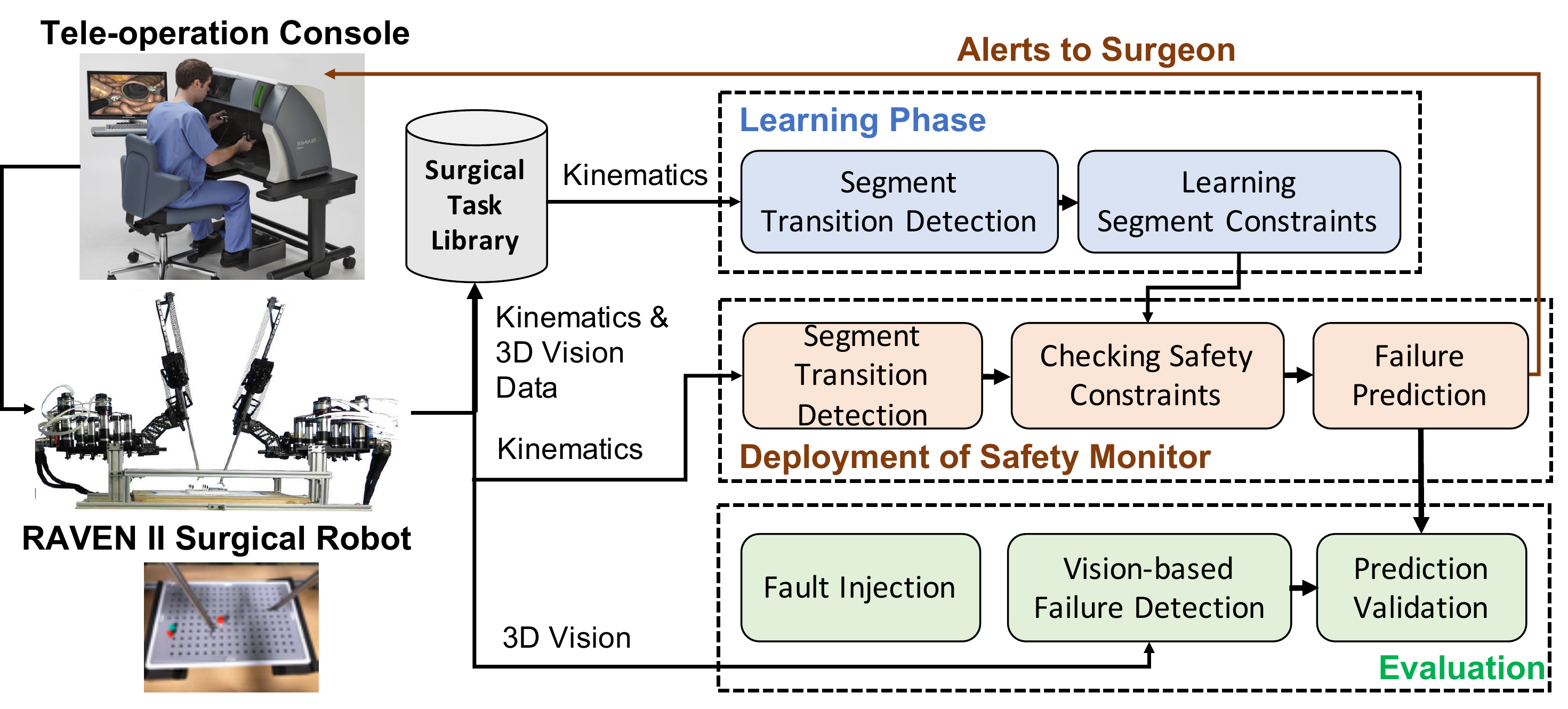}
    \caption{Context-Aware Safety Monitoring Pipeline}
    \label{fig:pipeline}
\vspace{-1.5em}
\end{center}
\end{figure*}

\section{Related Work}
Previous works have explored automated surgical task segmentation and skill evaluation in robot-assisted surgery. However, only a few previous works focused on ensuring safety and security. Here, we describe the most relevant work for our  monitoring system on surgical task segmentation, surgical skill evaluation, and safety in robotics.

\shortsection{Surgical Task Segmentation} Several previous works have also considered the problem of segmenting a surgical procedures into subtasks. Lalys et al.~\cite{lalys2011application} focused on cues obtained from images to identify the context  for cataract  surgery. The authors use color histogram intersection and Scale Invariant Feature Transform/Speeded-Up Robust Features (SIFT/SURF) for detecting the context based on the texture of objects. They use an AdaBoost classifier for detecting specific instruments in a sub window. Our segmentation approach is based on kinematics, 
but could be extended to detecting the context based on texture of objects using their approach. 
Krishnan et al.~\cite{krishnan2017transition} use a Dirichlet Prior to infer the number of possible clusters in the trajectory, assuming a categorical distribution. They deploy a hierarchical Gaussian Mixture Model (GMM) to cluster over kinematics and find the possible transitions, followed by at least another layer of clustering using spatial information to detect any transitions in between the ones already found. Finally, there is clustering with respect to time, to only include those transitions which are observed consistently across multiple demonstrations to decrease spurious transitions. The process is applied across trajectories, with clusters that have a likelihood of an arbitrary percentage, say 70\%, actually chosen. Our approach for detecting transitions draws inspiration from the GMM approach, however, we assume the number of subtasks is known a priori, and we only use the kinematics features instead of using both kinematics and vision (this is motivated by the possible inadequacies of vision feedback and our desire to use the vision data for independent evaluation). Other works have also looked at surgical task segmentation such as Fard et al.~\cite{fard2017soft}, who propose a soft-boundary unsupervised gesture segmentation using an unsupervised bottom-up approach. Their method starts from the finest possible data of the surgery and merges it with temporally similar data until it meets a criteria.   

\shortsection{Surgical Skill Evaluation} Several projects have focused on monitoring trajectories for surgical skill evaluation, with the goal of making surgery safer.  Nisky et al.\ \cite{nisky2014uncontrolled} demonstrate that experts' movements were more accurate, faster, and smoother than those of novices, especially in tele-operation. However, the JIGSAWS~\cite{gao2014jhu} data shows that experts do not always score highest; indeed, in some instances, experts had the lowest reported scores. Brown et al.\ \cite{brown2017using} used an ensemble learner, composed of different regression models, as well as a classification learner, to evaluate skill against manually annotated human scores. Their domain for evaluation included depth perception, force sensitivity and robot control, which are critical factors for evaluating the safety of tasks, with depth perception directly related to optimal point of view while robot control being related to the Cartesian and grasper angle of the end-effector. While skill evaluation techniques aim to make surgery safer by providing offline feedback, our work focuses on providing feedback to assist surgeons during surgery. 

\shortsection{Safety in Robotics} Most prior work on robotics safety focused on safety of human-robot interactions. For example, Ikuta et al.\ \cite{ikuta2003safety} focus on the control strategy of the robots in human-care applications. Lasota et al.~\cite{lasota2017survey} studied the contact force as a possible pre-collision parameter in human robot interactions. This can also be applied to surgical robotics where the forces applied by end-effectors are crucial parameters for safety. Other parameters considered include the grasper angle and Cartesian position, both of which affect the safety property of the segment and consequently its integrity. Our previous work on safety monitoring in robotic surgery \cite{alemzadeh2016targeted} proposed an anomaly detection technique based on real-time simulation of surgical robot dynamic behavior and preemptive detection of safety hazards such as abrupt jumps of end-effectors, but did not consider the underlying context or the specific surgical sub-tasks for detection.

\section{Context-Aware Safety Monitoring}\label{sec:context}
This section explains why it is important to consider context in surgical safety properties, and presents our proposed context-aware monitoring system for detection of safety constraint violations before they lead to adverse events. Figure~\ref{fig:pipeline} shows the overall monitoring pipeline, which is divided into two phases: \textit{Learning} and \textit{Deployment} (\emph{Evaluation}, also depicted in the figure, is discussed in Section~\ref{sec:evaluation}). In the learning phase, the system takes a set of correct trajectories as input, identifies subtask transitions, and infers local constraints specific to each subtask. 
We choose constraints over other approaches such as signature-verification  because we want to find properties that are consistent across all surgeons doing the same subtask. While there may be different signatures, the constraints should be the same for all safe operations. In the deployment phase, the safety monitor observes the  trajectories, recognizes subtask transitions, and applies the subtask-specific constraints to detect any potential safety violations. 
\subsection{Context in Surgery}
\begin {figure}[hbp!]
\vspace{-1em}
\begin{center}
    \includegraphics[width=0.7\linewidth]{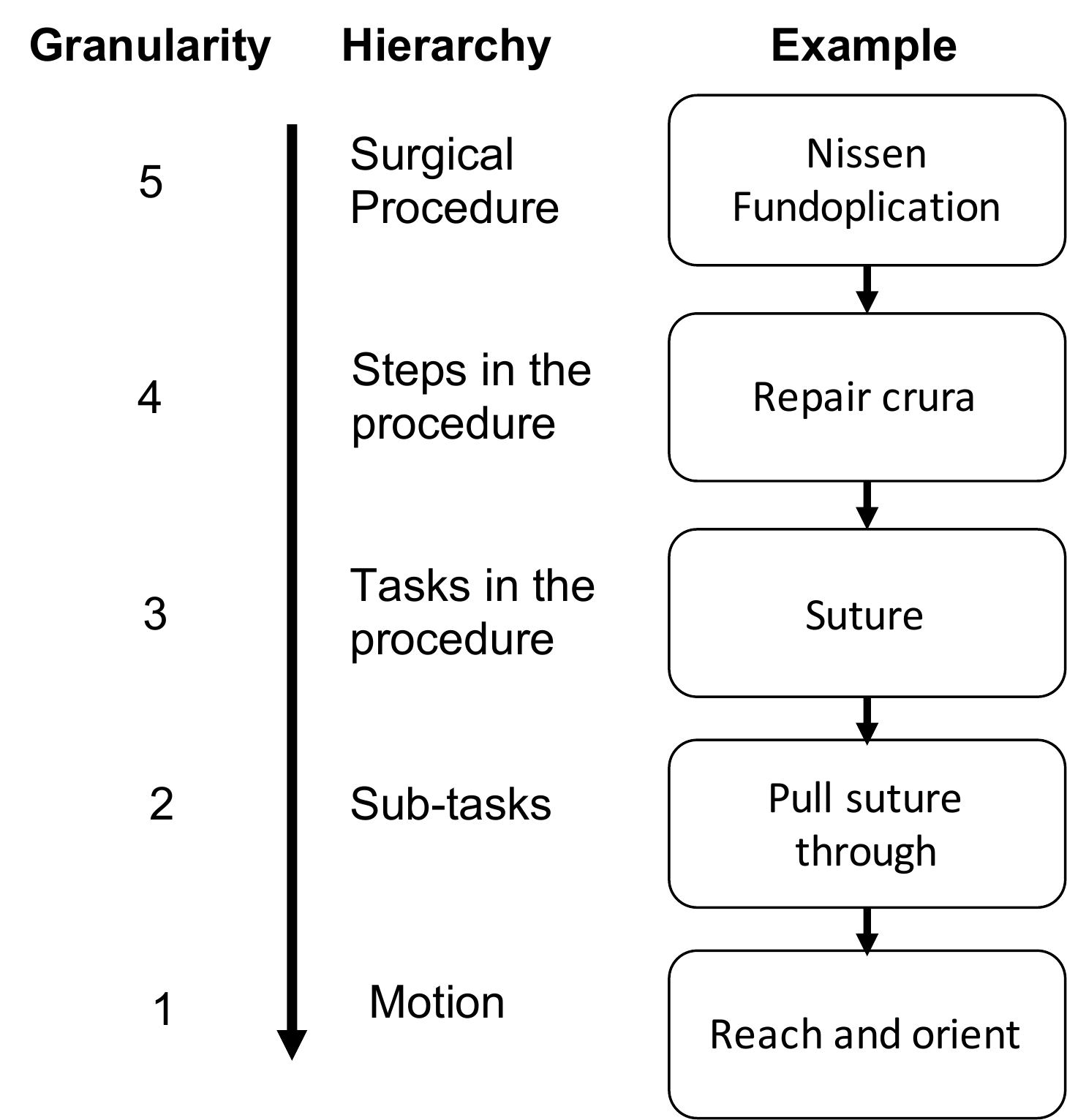}   
    \caption{\label{fig:hierarchy} Hierarchies in Surgical Procedures (adopted from \cite{neumuth2011modeling})}
\end{center}
\vspace{-1em}
\end{figure}

Context in surgical procedures can be organized into a hierarchy, starting from the surgical procedure that is being executed to the steps in the procedure to finally the specific motions of the robot (Figure~\ref{fig:hierarchy}).  Within a specific procedure (e.g., Nissen Fundoplication) or a surgical task (e.g., suturing), the change in context happens in the temporal domain as a result of the change of the surgeon's gesture or the position and orientation of the instruments end-effectors, leading to the corresponding change in the subtask (e.g., pull suture through). In our work, we assume that the task itself does not change until it has reached its end point, i.e., if the task in question is debridement (the removal of unhealthy tissue from a wound to promote healing), it doesn't abruptly change to another task, say knot tying, unless it has completely finished. As an example, Table~\ref{table:subtasks} shows the subtasks in a debridement task. Throughout the paper, we use the terms surgeme, subtask, and segment interchangeably.

\subsection{Task Segmentation}
The first stage for our context-aware safety monitoring framework is to detect transitions between subtasks. For this, we implemented a hierarchical clustering approach, by modeling the trajectory of a given surgical task (e.g., debridement) as a multimodal Dirichlet Distribution of $c$ clusters, with $c < N$, where $N$ is the number of samples in the surgical task. We adopted an approach similar to Krishnan et al.~\cite{krishnan2017transition} where three layers of clustering are used to detect transitions, based on the Cartesian position of the end-effectors, spatial changes (if any), and temporal changes. They particularly needed the spatial data to infer the grasper opening and closing or movement of the insertion tool based on the current segment, which was not directly  obvious from the kinematics data. We only use one layer of clustering based on the kinematics features from the Raven II with all the transitions of the task being detected within the single layer. We segment using joint values because they provide more precise features for distinguishing between gestures and surgical segments, compared to just using Cartesian or Cartesian and orientation features. For over-segmentation, we prune by using reference transitions which are the percentage in the timeline where the transitions are most likely to occur.  

\begin {figure*}[tb]
    
    \begin{subfigure}[b]{0.5\textwidth}
        \centering
        \includegraphics[width=1\linewidth]{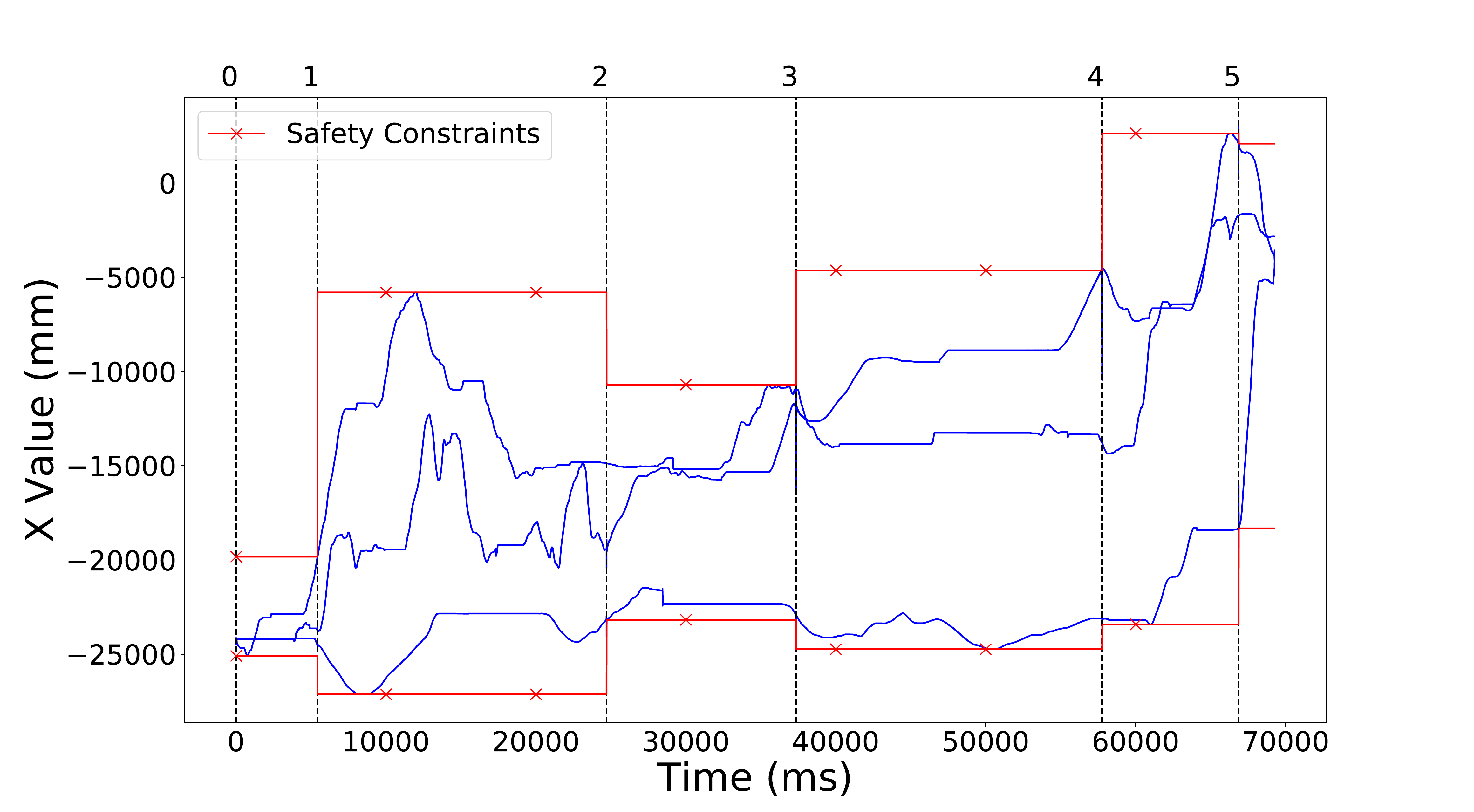}
        \caption{Cartesian X Position Trajectories along with Safety Constraints}
    \end{subfigure}%
    \begin{subfigure}[b]{0.5\textwidth}
        \centering
        \includegraphics[width=1\linewidth]{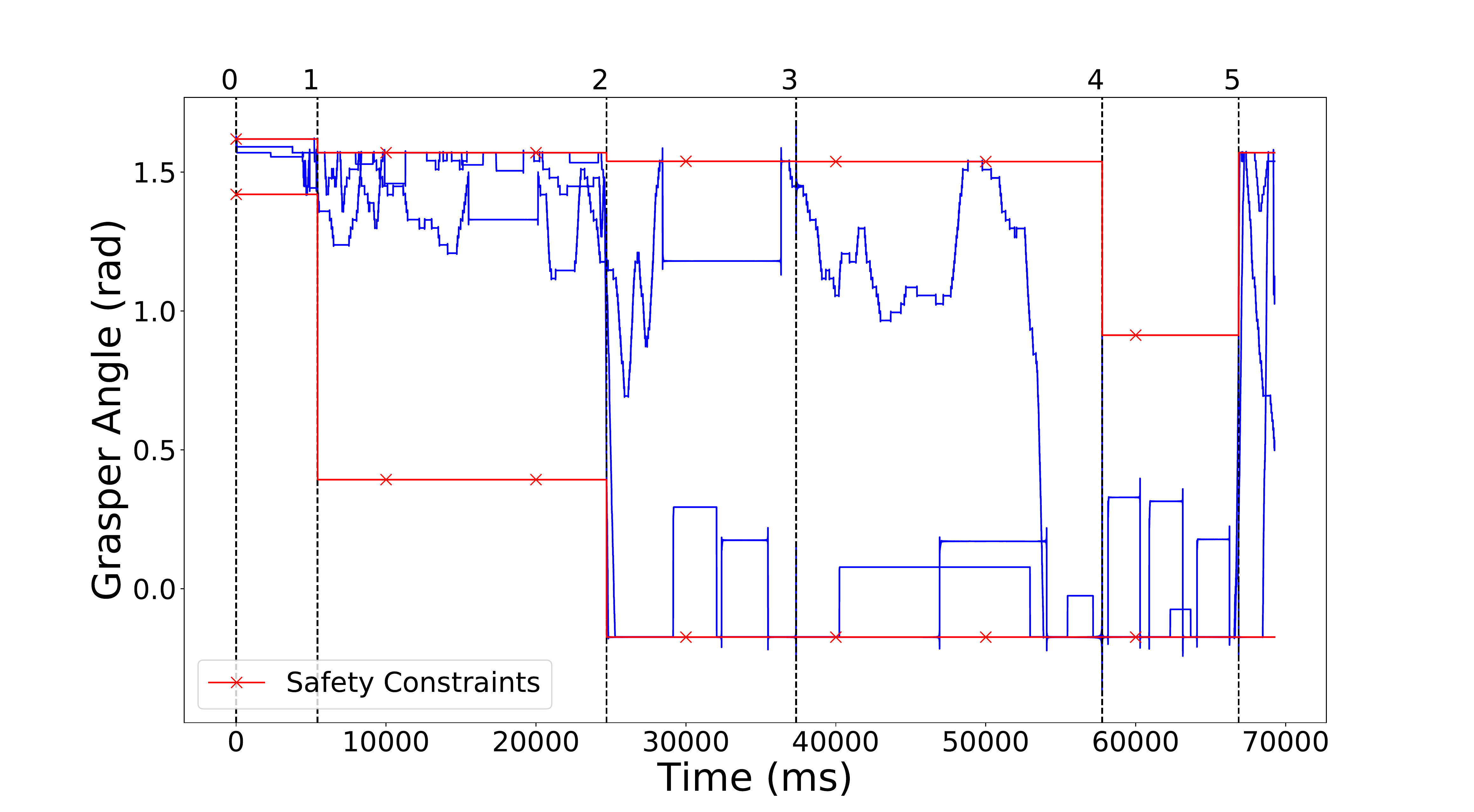}
        \caption{Grasper Angle Trajectories along with Safety Constraints}
    \end{subfigure}%
    \caption{Checking of Safety Constraints in Three Different Trajectories of Debridement Task.}
    \label{Fig:Constraint Difference}
    \vspace{-1em}
\end{figure*}

Krishnan et al.~\cite{krishnan2017transition} use visual features in their hierarchical clustering to deduce the state of the graspers which provide a more accurate estimate of the grasper pose compared to the values inferred from inverse kinematics. In this work, we do not incorporate visual features in the segmentation since we use them for detection of failures and evaluation of our proposed monitoring system and want it to provide an independent input from our kinematics based approach. 


\subsection{Safety Constraints}

Segment-specific constraints can be learned from the past data on safe trajectories taken for the same surgical tasks. Typical logic-based schemes associated with Fault Detection and Isolation (FDI) algorithms depend and rely on a set of constant or fixed constraints \cite{perhinschi2006adaptive}. For a fixed constraint, the maxima and minima should be tight enough to reliably detect impending failures (low false negative rate), without being violated on safe trajectories (low false positive rate) due to noise or inconsistencies in the execution of the task. Targeting these goals, we use locally-linear adaptive constraints for each segment. These are obtained using a learning-from-demonstration approach by analyzing trajectories from tele-operation data on the Raven II. Figure~\ref{Fig:Constraint Difference} shows an example of the safety constraints for the Cartesian position and grasper angle for the debridement task. These safety constraints are derived from the state-space of all the correct trajectories. For every correct segment, we take the highest and lowest values of Cartesian position and grasper angle. We propose that exceeding these constraints could lead to adverse events.   

Each segment of a surgery must be intrinsically safe to ensure the integrity of the entire procedure. Since each subtask has its own distinguishing properties, it is important to have distinct safety constraints for each subtask.  Based on the data from the JIGSAWS database~\cite{gao2014jhu}, for dry lab suturing in particular, our observation from manual analysis and review of JIGSAW's data showed that orientation of the needle driver was crucial for successful needle insertion and exit. On the other hand, for a task such as knot tying, the force with which the knot was tightened proved to be significant, since any higher or lower force could either make the knot too loose or snap the string. Hence, a crucial parameter that could discriminate between safe and unsafe actions is the applied force. Lastly, the Cartesian coordinates of the end-effectors along with the grasper angle are important parameters to be considered when determining safety of motions in a given subtask. 
In this paper, we use the Cartesian values of the end-effectors along with the grasper angle as the main parameters to monitor safety in each segment of a surgical task. Depending on the nature of the segment, these parameters may vary but can be integrated to our pipeline.



\subsection{Fault Model and Failure Modes}
We consider human errors, mechanical faults of the robot, and network vulnerabilities as the possible faults that might affect the performance of the surgical task. Human errors might happen due to lapses of concentration or lack of training~\cite{sarker2005errors}. Previous works \cite{alemzadeh2016targeted, alemzadeh2014systems, bonaci2015make} have documented the potential for software and network faults and vulnerabilities causing unwanted robotic motions. To model such errors, we use an additive fault injection approach where errors are added to the safety parameters of a correct trajectory for a certain amount of time. The target parameters are the Cartesian position of the end-effector (sudden jumps) and the grasper angle (closing or opening of the grasper). Table \ref{table:common_errors} summarizes common types of human errors reported in the literature. In our experiments, we inject faults designed to roughly mimic such common errors. These fault injections could also apply for mechanical faults in the robot, where the device is producing unwanted movements, like sudden jumps or the surgeon's movement is not followed, such as the grasper not opening. 

Table \ref{table:failure_modes} lists the typical failure modes that were observed due to the fault injections in our experiments along with the segment that was specific to each failure. Although our reported failures may seem trivial due to our experimental dry-lab setup, they could cause serious harm in real surgical scenarios. For example, dropping a block in a dry-lab setting could simulate dropping a tiny needle or cancerous tissue during a procedure. The sudden jumps in dry-lab simulation could result in tissue or organ damage in a surgery.

\begin{table}[tb]
\begin{center}
\caption{Common Errors in Surgery}
\label{table:common_errors}
\setlength\tabcolsep{5pt}
\begin{tabular} {|c|c|c|}
 \hline
\textbf{Error} & \textbf{Task} & \textbf{Reason} \\
\hline
Unintentional release & Laparoscopic & Grasper angle  \\ 
of gall bladder~\cite{eubanks1999objective} & cholecystectomies & higher than usual 
\\
\hline
Droping needle/slipping & Suturing of & Wrong grasper  \\
 of knot after tying~\cite{elhage2015assessment} & an anastomosis & 
 angle/orientation \\ 
\hline
Too much force tearing & Laparoscopic & Wrong scale factor \\
gall bladder~\cite{joice1998errors} & cholecystectomies & / grip too tight \\
\hline
Wrong site injury~\cite{kwaan2006incidence} & General &
Wrong position  \\ 
\hline
\end{tabular}
\end{center}
\vspace{-1 em}
\end{table}

\section{Experimental Evaluation}\label{sec:evaluation}

Our experiments used the debridement task which removes dead or harmful tissue from the body. This task contains segments where the robot end-effector is in direct contact with the tissue, which could potentially lead to unsafe scenarios. For simulating debridement in a dry-lab setting, we used the block-transfer setup, with the blocks mimicking the tissues in real surgery. As shown in Figure~\ref{fig:rq2_2}, the debridement task can be segmented into six distinct segments or subtasks.  

\subsection{Experimental Setup}
Our experiments used a Raven II robot (Applied Dexterity, Inc.), integrated with a dVTrainer tele-operation console (Mimic Technologies, Inc.). We used a ZED Mini camera (Stereolabs, Inc.) for 3D vision for both tele-operation and recording. The kinematics data was logged using the native ROS API \cite{quigley2009ros} at 1000 packets per second and the vision data was recorded with the ZED Mini SDK at 30 frames per second (fps). For evaluation purposes, the kinematics data was synchronized with the vision data by using their corresponding timestamps in epoch time and converting them to datetime.

We logged kinematics and vision data from correct trajectories recorded from human subjects operating on the Raven II robot. The subjects in our experiments did not have any experience with the surgical console and were only given 30 minutes to familiarize themselves with the workspace and the console. In total, we collected 10 fault-free trajectories of debridement task from 5 subjects. This data was divided into two separate sets for training (made up of fault-free trajectories) and testing (made up of faulty trajectories generated using fault injection). We used the training data to learn the segment-specific safety constraints from the kinematics data. We used the testing data for fault injection experiments and evaluation of the safety monitor using vision. We labeled the transitions in segments from the video data. 

\subsection{Fault Injection}
\begin{figure} [tb]
    \captionsetup[subfigure]{justification=centering}
    \centering
    \begin{subfigure}{.25\textwidth}
    \centering
    \includegraphics[width=\linewidth , trim=40 10 50 50, clip]{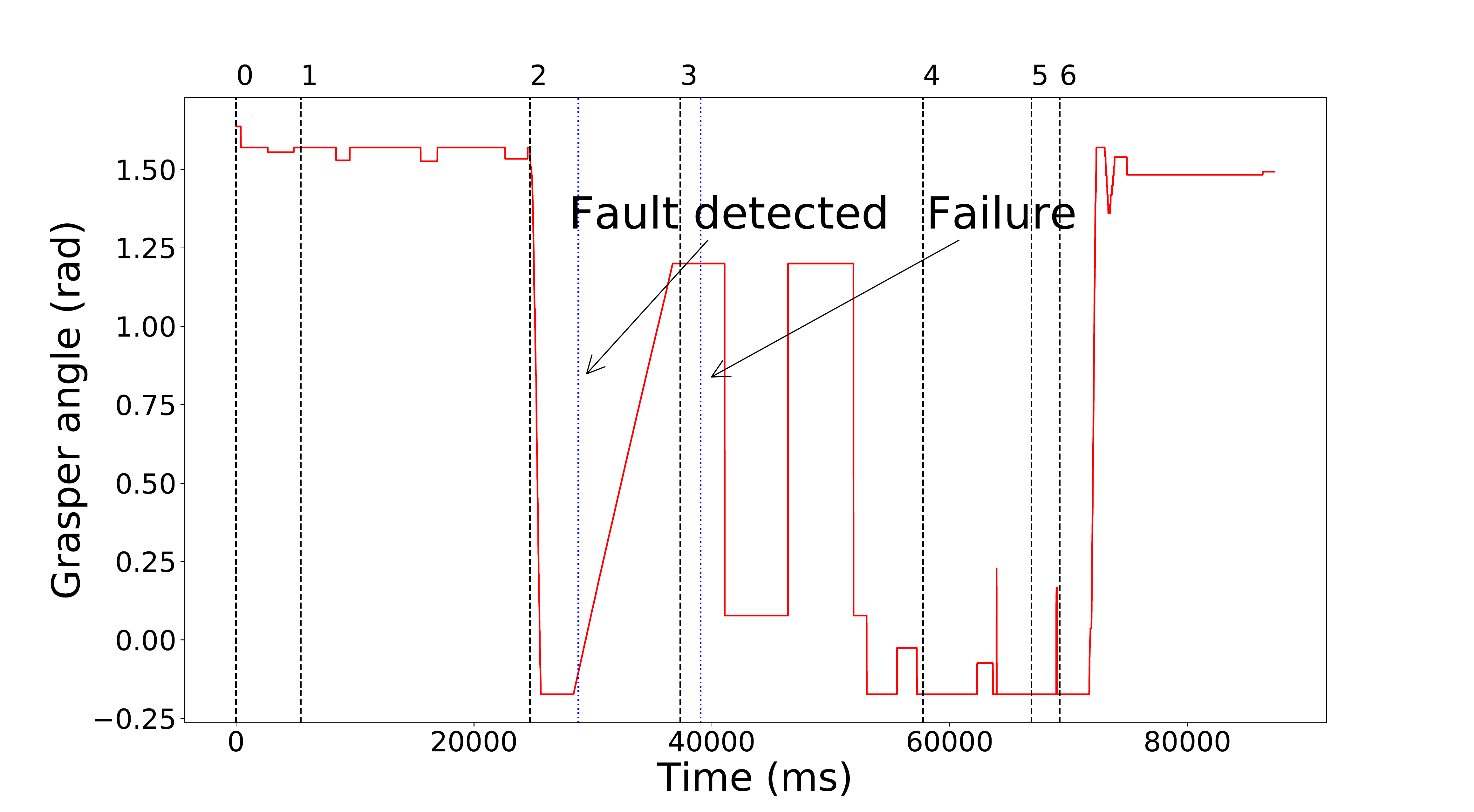}
    \caption{\centering Grasper Angle}
    \label{Fig:fault injection grasper}
    \end{subfigure}%
    \begin{subfigure}{.25\textwidth}
    \centering
    \includegraphics[width=\linewidth , trim=30 10 50 50, clip ]{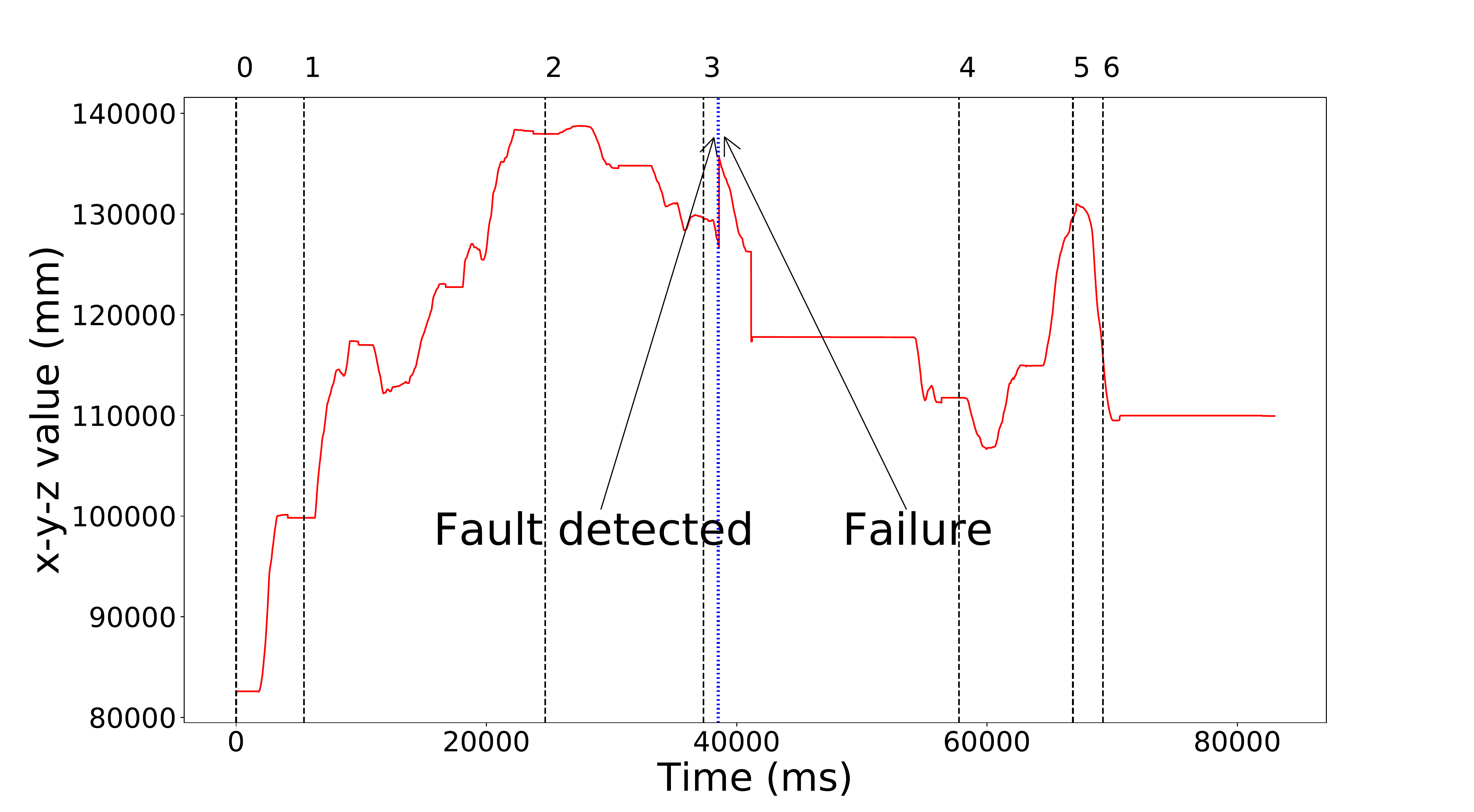}
    \caption{\centering Cartesian Positions}
    \label{Fig:fault injection cartesian}
    \end{subfigure}
    \caption{Impact of Faults on Correct Trajectories}
    \label{Fig:Fault injections}
\end{figure}

We injected faults to the values of the safety parameters, to test whether they lead to any failures. For this, we overwrote the kinematics data and sent the faulty trajectory packets using a Master Console Emulator~\cite{alemzadeh2016targeted} to the robot control software. This allowed us to repeat the same trajectory or to perturb only certain segments of the trajectory while the rest of data remained the same. Figure~\ref{Fig:Fault injections}
shows the results of a fault injection experiment targeting the Cartesian positions and Grasper angles. We perturb the end-effector's position by forcing it to go to a randomly selected value ranging from 300 to 65000 mm and grasper angle by forcing it to go to a randomly-selected targeted value between 0.1 and 1.4 radians for different duration, with some cases involving corruption across segments and other cases involving just a portion of a segment. 

\subsection{Failure Analysis} 
We used vision data as an independent indicator for automated detection of failures and used it as the ground-truth for evaluation of our safety monitoring system. Table~\ref{table:failure_modes} summarizes the types of failures we considered, all of which were detected automatically using the vision. Automated detection of failures using vision not only enables more precise identification of the time of failure, but also labeling of larger data sets collected from surgery for more comprehensive evaluation of the safety monitor or even offline skill evaluation. 

\begin{table}[bt]
\centering
\begin{center}
\caption{Observed Failure Modes}
\label{table:failure_modes}
\begin{tabular} {|c|c|c|}
 \hline
\textbf{Failure} & \textbf{Cause} & \textbf{Segment} \\
\hline
Unintentional release & Grasper angle too high & 4 \\
& or Wrong scale factor & \\
\hline
Failure to dropoff & Grasper angle too low  & 5 \\
\hline
Hitting an obstacle  & Wrong Cartesian position & all \\
 & Wrong scale factor & \\
\hline
\end{tabular}
\end{center}
\vspace{-2em}
\end{table}

\begin{figure}[bt]
    \centering
    \begin{subfigure}{.25\textwidth}
    \centering
    \includegraphics[width=\linewidth, trim=50 80 60 80, clip ]{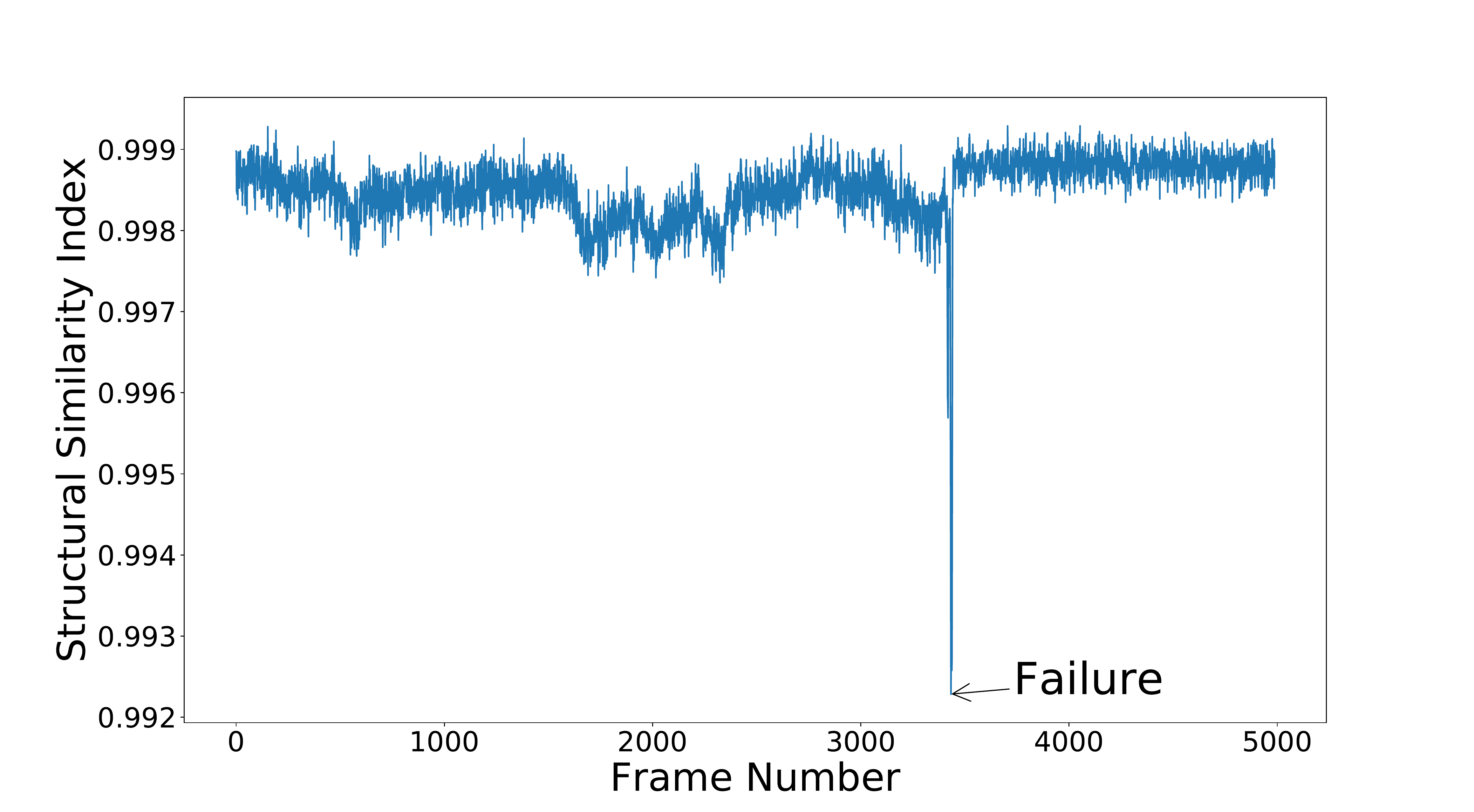}
    \caption{Structural Similarity Index}
    \label{fig:sub1}
    \end{subfigure}%
    \begin{subfigure}{.25\textwidth}
    \centering
    \includegraphics[width=\linewidth, trim=50 80 60 80, clip]{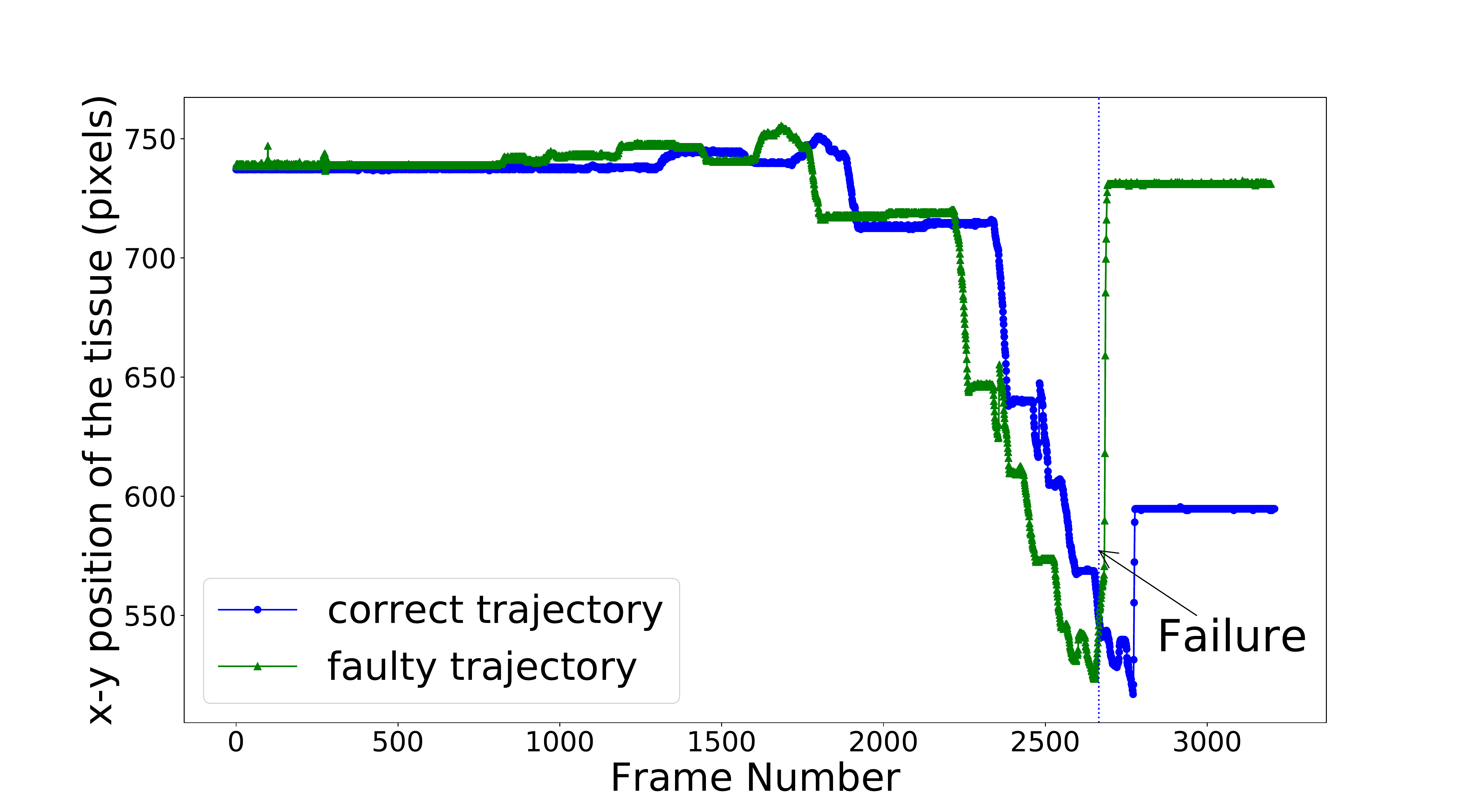}
    \caption{Dynamic Time Warping}
    \label{fig:sub2}
    \end{subfigure}
    \caption{Automated Failure Analysis using Vision.}
    \label{fig:test}
\end{figure}

\shortsection{Structural Similarity Index}
One approach we used to detect failures from the vision data was based on the \emph{structural similarity index}~\cite{wang2004image}. The structural similarity index finds the similarity measurement of two images (in our case, adjacent frames in the video) with respect to luminance, contrast and structure. 
Figure~\ref{fig:sub1} shows an example where the block was unintentionally dropped at frame 3434. In this example, the local constraint for grasper angle was exceeded for multiple segments before the failure. To localize the exact packet in the kinematics where the failure occurred, we compare the recorded video frames of the trajectory to find where the difference between adjacent frames is greatest. As seen in the Figure~\ref{fig:sub1}, there is a sharp peak at the point of failure, which would not have occurred if there were no failures. We then match the timestamp of the frame with the corresponding packet in the kinematics data to localize the failure.

\shortsection{Dynamic Time Warping}
To detect cases where the block is not dropped at the right position or not picked at all, the image similarity metric is insufficient since it contains no information about the position of the block. For these cases, we first threshold the image to detect the block and then apply contour detection to find its center. We compare the position of the block as detected with a fault free trajectory and apply dynamic time warping to find the highest deviation between corresponding frames (Figure~\ref{fig:sub2}). The highest deviation is likely to indicate the point where a failure occurs.

\begin{table}[tb]
\caption{Transition Detection Accuracy}
\label{Table: Jaccard Simmilarity for TSC}
\centering
\begin{tabular} {|c|c|c|}
\cline{2-3}
\multicolumn{1}{c|}{} & \textbf{Task (Dataset)} & \textbf{Jaccard Similarity}  \\
\hline
\multirow{2}{*}{TSC \cite{krishnan2017transition}}&Needle Passing (JIGSAWS) & 83\% \\ &Suturing (JIGSAWS) &73\% \\
\hline
\multirow{2}{*}{This Work} 
&Knot Tying (JIGSAWS) &68\% \\
&Debridement (RAVEN) &75\% \\
\hline
\end{tabular}
\end{table}

\begin{table}[bt]
\begin{center}
\caption{Average Error in Subtask Transition Detection}\label{table:subtasks}
\label{table: jitter}
\centering
\begin{tabular} {|c|l|c|}
 \hline
\textbf{Subtask}  & \multicolumn{1}{|c|}{\textbf{Name}} & \textbf{Avg. Error ($\Delta t$ in frames)} \\
\hline
0 & Start & -56 \\
\hline
1 & Moving to the block & 76 \\
\hline
2 & Grabbing the block & -69 \\
\hline
3 & Moving up & -30 \\
\hline
4 & Moving to the receptacle & -10 \\
\hline
5 & Dropping the block & -3 \\
\hline
6 & End & 0 \\
\hline
\end{tabular}
\vspace{-2em}
\end{center}
\end{table}

\subsection{Experimental Results} \label{sec:experiments}
First, we report on the accuracy of the segment transition detection, since the safety constraints applied depend on the predicted subtask. Then, we present results to evaluate our overall goal of accurately anticipating safety violations. 
\par
\shortsection{Segment Transition Detection}
We use the \textit{Jaccard Similarity} to measure the percentage overlap between the predicted transitions and the ground truth transitions. We also report the jitter, $\Delta t$, which measures how early or late the transitions are predicted. For the knot tying task~\cite{gao2014jhu}, our segmentation has an average of 68.0\% overlap with the ground truth segments of knot tying based on only Cartesian and grasper values, whereas the percentage overlap is lower for more complex tasks such as suturing where the use of visual features is particularly needed. We used two layers of clustering based on only kinematics. The accuracy Table \ref{Table: Jaccard Simmilarity for TSC}  can be improved by using information from vision data, followed by loop compaction and pruning, similar to Krishnan et al.\ \cite{krishnan2017transition}. For debridement, the average accuracy is 75.4\% . For jitter (see Table \ref{table: jitter}), the maximum lateness between the predicted and the ground truth transition was 76 frames, while the minimum was -3 frames. 

\shortsection{Safety Violation Detection}
Table \ref{table:conf_cart} shows the results of our fault injections for perturbing the Cartesian position and grasper angle values respectively, to induce sudden jumps that could cause unintentional release or pickup failure. It counts instances where injection of faults led to violations of the safety constraints and resulted in a failure prediction. To evaluate the efficacy of the detector, we used vision data to determine whether or not a failure actually occurred. We consider it a successful prediction if a failure prediction is followed by a failure, and a false positive if no failure occurs following a failure prediction. Sometimes an injected fault does not result in a violation of any safety constraint, so no failure was predicted. We consider it a false negative if a failure occurs that was not preceded by a constraint violation.
\par

The timing of the failures was confirmed using the vision data and overlaid with the kinematics data to localize the exact packet where the failure occurred. We measured the average time to react, defined as the time between the detection of safety violation by the monitor and the actual occurrence of a failure as shown in Table \ref{table:conf_cart}. Our experiments confirm that failures usually happen after prolonged violation of local constraints (see Figure \ref{Fig:fault injection grasper}) and there is usually enough time between the detection of safety violation and the actual failure (last column of Table \ref{table:conf_cart}). These results provide optimism that alerting a surgical team on violations of safety constraints could be useful in preventing adverse events. 

Further analysis of the fault injection results indicates that exceeding of the grasper angle by a small margin does not result in dropping the block, while grasper angles above 0.8 radians led to unintentional release of the block. Grasper angles below 0.8 radians led to failure to drop the block at the right time and place. Our experiments also indicate that the parameters such as the time (in a segment of surgical task) at which the faults are injected and the duration for which they are injected play a role in causing the violation of safety constraints. More in-depth analysis of the impact of different fault types and locations is the subject of future work.\par

\begin{table}[tb]
\centering
\begin{center}
\caption{Safety Violation Detection Results}
\label{table:conf_cart}
\begin{tabular} {|c|c|c|c|c|}
     \hline
     \textbf{Simulated}& \textbf{Predicted} & \multicolumn{2}{c|}{\textbf{Actual Outcome}} & \textbf{Time to}\\
     \textbf{Failure} & \textbf{Outcome} & \multicolumn{2}{c|}{\textbf{(Grandtruth)}} & \textbf{React} \\
     \textbf{Scenario} & \textbf{(Safety Monitor)} & \textbf{Failure} & \textbf{No Failure} &  \textbf{(seconds)} \\
    \hline
     \multirow{4}{*}{} & Failure (0 sd) & 10 & 7 & 1.7\\
    Sudden& No Failure (0 sd) & 0 & 3 & \\
    Jump& Failure (1 sd) & 0 & 0 & -\\
    & No Failure (1 sd) & 10 & 10 &  \\
    \hline
    \multirow{4}{*}{} & Failure (0 sd) & 6 & 2 & 14.4\\
    Block& No Failure (0 sd) & 0 & 8 &  \\
    Drop& Failure (1 sd) & 6 & 1 & 11.4\\
    & No Failure(1 sd) & 0 & 9   & \\
    \hline
\end{tabular}
\vspace{-2em}
\end{center}
\end{table}

\section{Conclusion}

We presented preliminary results from our experiments developing a context-aware safety monitoring system that could alert the surgeon of the impending safety-critical events. Our results provide encouraging evidence that there is an opportunity for using violations of subtask-specific safety constraints to warn surgeons early enough to prevent impending adverse events. Our results so far are limited to one task, debridement, and based on a small number of trials with non-experts using a surgical robot in a dry-lab environment. Many challenges remain before such an approach could be used in practice, including validating the effectiveness of safety constraints in a realistic environment and across a range of operators, determining a safe and effective way to convey alerts to the surgical team, and understanding the impact of any surgeon-assistance features on regulation of robotic devices. Nevertheless, we are optimistic that the data available from a surgical robot system can be used during procedures to enhance safety.





\section*{Code Availability}

The code for the safety monitor and the fault injection experiments is available at the following repository: {\sloppy
https://github.com/UVA-DSA/ContextMonitor}.

\section*{Acknowledgments}

This work was partially supported by a grant from the National Science Foundation (1804603) and a research innovation award from the School of Engineering and Applied Science (SEAS) at the University of Virginia.





\bibliographystyle{IEEEtran}
\bibliography{surgical_context}
%



\end{document}